\documentclass[lettersize,journal]{IEEEtran}
\usepackage{amsmath,amsfonts}
\usepackage{algorithm}
\usepackage{array}
\usepackage[caption=false,font=normalsize,labelfont=sf,textfont=sf]{subfig}
\usepackage{textcomp}
\usepackage{stfloats}
\usepackage{url}
\usepackage{verbatim}
\usepackage{graphicx}
\usepackage{cite}
\usepackage{hyperref}
\usepackage{tabu}                      
\usepackage{booktabs}                  
\usepackage{amsmath}
\usepackage{amssymb}
\usepackage{bm}
\usepackage{multirow}
\usepackage{makecell}
\usepackage{comment}
\usepackage{mathptmx}                  
\usepackage{amsmath,amssymb,amsfonts}%
\usepackage{amsthm}%
\usepackage{mathrsfs}%
\usepackage{manyfoot}%
\usepackage{algpseudocode}%
\usepackage{listings}%
\usepackage{bm}%
\usepackage[numbers,sort&compress]{natbib}
\usepackage{enumerate}
\usepackage{lettrine}
\usepackage{color}

\renewcommand{\emph}[1]{\textbf{#1}}

\newcommand{\etal}{\textit{et al.}}
\newcommand{\ie}{i.e.}

\newcommand{\fref}[1]{{Fig.~\ref{#1}}}
\newcommand{\tref}[1]{{Table~\ref{#1}}}
\newcommand{\sref}[1]{Section~\ref{#1}}

\begin{document}

\title{Disturbance-Free Surgical Video Generation \\
from Multi-Camera Shadowless Lamps for Open Surgery}

\author{%
Yuna Kato, 
Shohei Mori,~\IEEEmembership{Member,~IEEE},
Hideo Saito,~\IEEEmembership{Senior Member,~IEEE}, \\
Yoshifumi Takatsume,
Hiroki Kajita,
Mariko Isogawa,~\IEEEmembership{Member,~IEEE}
}

\markboth{}%
{Shell \MakeLowercase{\textit{et al.}}: A Sample Article Using IEEEtran.cls for IEEE Journals}


\maketitle

\begin{abstract}
Video recordings of open surgeries are greatly required for education and research purposes. However, capturing unobstructed videos is challenging since surgeons frequently block the camera field of view. To avoid occlusion, the positions and angles of the camera must be frequently adjusted, which is highly labor-intensive. Prior work has addressed this issue by installing multiple cameras on a shadowless lamp and arranging them to fully surround the surgical area. This setup increases the chances of some cameras capturing an unobstructed view. However, manual image alignment is needed in post-processing since camera configurations change every time surgeons move the lamp for optimal lighting. This paper aims to fully automate this alignment task. The proposed method identifies frames in which the lighting system moves, realigns them, and selects the camera with the least occlusion to generate a video that consistently presents the surgical field from a fixed perspective. A user study involving surgeons demonstrated that videos generated by our method were superior to those produced by conventional methods in terms of the ease of confirming the surgical area and the comfort during video viewing. Additionally, our approach showed improvements in video quality over existing techniques. Furthermore, we implemented several synthesis options for the proposed view-synthesis method and conducted a user study to assess surgeons' preferences for each option.
\end{abstract}

\begin{IEEEkeywords}
Open surgery, video synthesis, multi-view camera calibration, event detection, camera selection
\end{IEEEkeywords}

\section{Introduction}\label{sec:intro}
\IEEEPARstart{S}{urgical} videos can provide objective records in addition to medical records. Such videos are used in various applications, including education, research, and information sharing~\cite{education, research, tasks}.
In endoscopic and robotic surgery, the surgical field is kept visible as the camera device replaces the surgeon's view for monitoring the operations~\cite{endoscopy}.
However, in open surgery, surgeons are in-situ, and cameras in the same room must observe the surgical field \cite{open_surgery}. This setup inevitably increases chances of obstructions in the camera field of view.

\begin{figure}[tb]
\begin{center}
\includegraphics[width=\hsize]{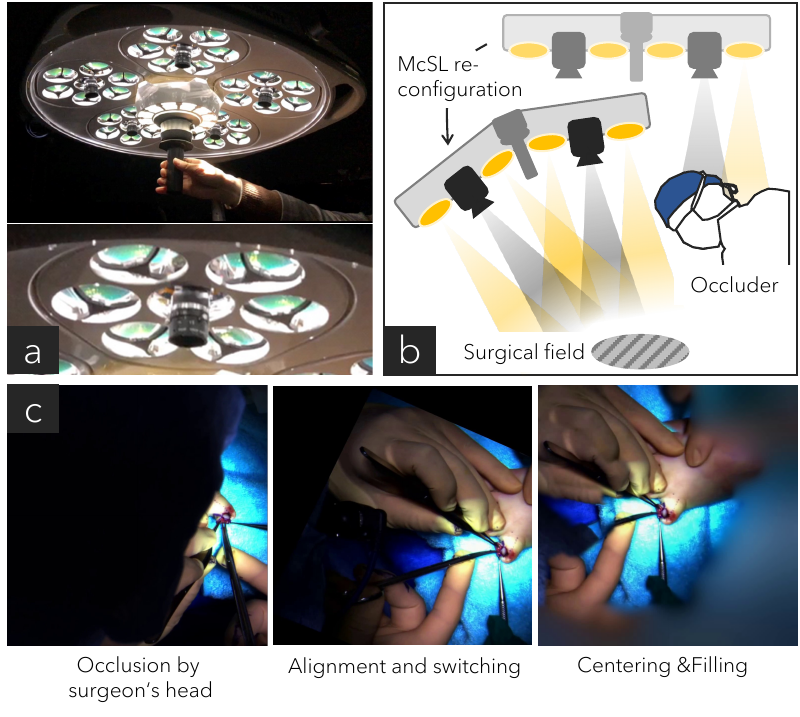}
\caption{Multi-camera shadowless lamp (McSL).
    (a) The system comprises five units, each with six light sources and a camera (bottom).
    (b) The location and focus are adjusted to the surgical field by moving the central knob, causing camera configurations to change over time. Occluded light sources are compensated by others in conventional shadowless lamps, and so are cameras in McSL.
    (c) To avoid occlusions in output videos, a previous approach \cite{Shimizu} switches an occluded camera to another with fewer occlusions. However, this results in rotated cameras over time. Our approach \textit{automatically} aligns frames, centers the surgical field for stabilized outcomes, and detects frames upon motion for reconfigurations. We further apply visual aids to re-center the view and inpaint missing pixels.
    }
\label{fig:mcsl}
\end{center}
\end{figure}

To overcome this issue, Kumar and Pal~\cite{Kumar} used a stationary arm with a camera at the tip.
This is potentially an additional load for the surgeons who set it up and may disturb the surgery.
Mounting a camera on the surgeon's head~\cite{Nair, head_mount1, head_mount2} is a more standard approach, while it suffers frequent and drastic motions.
Recent studies have installed cameras on shadowless lamps for solid and stable recordings. 
Byrd~\etal~\cite{Byrd} mounted a camera on a shadowless lamp while the surgeon's body could block it. 
To address this issue, Shimizu~\etal~\cite{Shimizu} developed a shadowless lamp with multiple cameras (i.e., multi-camera shadowless lamp; McSL) to ensure that at least one camera would observe the surgical field by switching the camera views (\figurename~\ref{fig:mcsl})~\cite{Shimizu,Hachiuma,Saito}.
However, since each camera captures videos of the surgical field from a different direction, it was necessary to manually align the videos in advance.
Obayashi~\etal~\cite{obayashi} developed a video player to manually seek and segment a video clip with no camera movement. This automatically performs camera calibration and homography matrix estimation from seeked video clips, while the automation remained as future work.
In addition, even though all of the above methods use McSL, they ignore the possibility that the surgeon may move the McSL. In practice, McSL do move, requiring a mechanism to detect movement of camera installed in the McSL for full automation.

Overall, the combination of McSL and its view-switching poses a new task that we address in this paper: Camera reconfiguration detection and multi-view image alignment for seamless view transitions with the least occlusion.
This paper extends our previous work~\cite{miccai}\footnote{project page: \color{magenta}https://github.com/isogawalab/SingleViewSurgicalVideo} presented at MICCAI 2023 by incorporating two key features into our video view synthesis process: view centering and pixel filling. These enhancements were rigorously evaluated through a professional review conducted by medical doctors.
Our contributions arethe following: 
\begin{itemize}
    \item We propose to synthesize a stabilized single-view video with minimized occlusions from a multi-camera system installed in a shadowless lamp.
    \item To this end, we propose algorithms to detect the start and end frames of moving McSL by measuring misalignment between the cameras and to find frames with less occluded surgical fields.
    \item We conducted a user study with surgeons and a quantitative video evaluation to assess the clarity of the surgical field and the stability of the video.
    \item We collected preferences from surgeons regarding various video synthesis options that provide design guidelines for video synthesis based on their feedback.
\end{itemize}

\section{Related work}
\label{sec:related}
\subsection{Surgical Video Recording and Processing}
For over a century, surgical education has been founded on Lord William Halsted's basic principle for training residents: \textit{``See one, do one, teach one''}~\cite{basic_principle}. However, with recent efforts to improve efficiency in the operating room and reduce medical errors, opportunities for residents to directly observe and learn in the surgical setting have become limited~\cite{changes}.

To address this issue, there is a growing movement to introduce tools to facilitate skill acquisition outside the operating room. For example, advanced simulation technologies like Virtual and Augmented Reality~\cite{vr,ar} allow residents to learn surgical skills and processes without any risks. Additionally, in precision surgeries, robot-assisted surgery enables practice in simulation mode and provides feedback~\cite{robot-assisted}. While these simulation-based trainings offer hands-on experiences, they are not yet readily accessible to individuals. Challenges lie in harmonizing virtual content with reality and their high costs.

Meanwhile, video-based education is accessible and effective to anyone, enabling surgical observation from anywhere. Some reports indicate that most residents and specialists currently use online video materials (e.g., on YouTube) to prepare for surgeries~\cite{video_education}. Moreover, surgical videos are valuable records for research and information sharing~\cite{research, tasks} beyond their educational use.

In recent years, minimally invasive surgery, such as endoscopic and laparoscopic procedures, has rapidly gained popularity in the field of surgery. Video recording has become easy and commonplace since these surgeries involve capturing operations with specialized cameras. The need for real-time observation of the surgical field in these areas has led to extensive research into the automation of camera control~\cite{camera_control}. On the other hand, open surgery remains the standard in many surgical specialties due to its advantages in handling cases involving large areas or requiring urgent response. However, there is no established video technology for recording in open surgery, making it challenging to capture high-quality videos. To address this issue, various cameras and placement methods have been proposed in the past.

The most common method for capturing videos in open surgery is to use a head-mounted camera~\cite{Nair, head_mount1, head_mount2}. By placing the camera on the surgeon's head, it records the view directly from the surgeon's perspective, which is considered to be the best viewpoint for the viewer. Higher quality has also been developed, such as a method for tracking the direction of the surgeon's line of sight~\cite{track_eye} and reducing blurring caused by small head's movements~\cite{stabilize}. However, since surgeons do not always keep their eyes on the surgical field due to interactions with other staffs or retrieving instruments, it is challenging to eliminate the effects of large head movements.

When using smart glasses type cameras, they offer the advantage of being lighter so more acceptable in the field compared to head-mounted cameras. However, their short battery life makes them unsuitable for long surgeries~\cite{smartglasses}. Additionally, similar to head-mounted cameras, they face challenges related to head movement.

Researches using arms or tripods for capturing~\cite{Kumar, tripod} have shown that these setups can stabilize the camera, eliminating issues with camera shake. However, challenges remain in capturing detailed areas due to issues such as overexposure. There is a research that have used laparoscopic cameras to capture high-resolution, high-contrast videos of the surgical field. However, it is challenging to position the camera without interfering with the operation, often requiring adjustments by the camera operator~\cite{laparoscopic}.

Researches using cameras mounted on shadowless lamp~\cite{Byrd} found that the camera's view was often obstructed by the surgeons' head or body, making it difficult to use continuously. Shimizu~\etal proposed a method to solve this issue without a camera operator~\cite{Shimizu}. In their research, multiple cameras were mounted on the shadowless lamp, and the optimal frames were automatically selected over time to address the occlusion issue. However, this approach introduced a new challenge: aligning multiple video frames, which required realignment whenever the shadowless lamp moved. So, our study proposes a method that automatically aligns the video frames, producing fully automated, unoccluded surgical videos that clearly show the surgical field.

\subsection{Video synthesis for occluded region}
Occlusion of target objects is a common occurrence in various situations. To address the image holes created in such cases, the concept of image inpainting was first introduced in \cite{first_inpainting}. Traditional methods include approaches that smoothly interpolate missing regions using surrounding pixel information~\cite{traditional_inpainting} and those that utilize similar patches within the image~\cite{patchmatch}. These methods can effectively blend missing areas with their surroundings, but they often struggle in challenging situations. For instance, they may fail when dealing with complex objects or when images lack similar textures because they do not understand the scene's context. In recent years, the use of deep neural networks has enabled handling more complex scenes~\cite{face, complex}. Moreover, by incorporating generative Adversarial Network (GAN)~\cite{gan} or combining with segmentation and object recognition models~\cite{sam}, it has become possible to generate more natural images.

In the case of video, directly applying image inpainting algorithms to each frame often introduces temporal artifacts, making it essential for video inpainting methods to incorporate temporal consistency constraints. The method proposed in \cite{space-time}, which extends patch-based inpainting approaches, searches for similar patches to fill missing regions while maintaining spatiotemporal coherence. However, this approach is computationally expensive and struggles in scenarios with prolonged occlusions. Method utilizing 3D neural networks~\cite{3dcnn} enables simultaneous processing of spatiotemporal information but suffers from limited receptive fields, reducing its effectiveness for distant regions. Some methods, like using optical flow to propagate inter-frame context~\cite{flow}, aim to overcome this limitation but struggle with large occluded areas due to inaccurate flow estimation. Other approaches, such as learning motion patterns specific to test videos through internal training~\cite{internal_training} or using transformers to model long-range dependencies~\cite{fuseformer}, improve video inpainting performance. However, these methods are computationally demanding, making practical use challenging.

In the medical field, inpainting techniques have been used to remove specular reflections from endoscope images~\cite{endoscope_reflection} and to remove metal artifacts from X-ray images~\cite{x-ray}, for example. However, these methods are typically limited to repairing small areas. In our system, occlusions like a head can cover large portions of the video, making it difficult to inpaint them seamlessly without introducing artifacts. Moreover, using generative methods, such as those favored by inpainting, in critical applications, such as surgical recording or diagnosis, demands caution. If these methods generate content based on similar cases, it may inaccurately represent the actual procedure, potentially causing serious issues later. Therefore, we would like to consider a method of synthesizing using viewpoints captured by other cameras, instead of smoothing holes in the image.

A related technique to inpainting is Diminished Reality (DR). Contrary to AR, it reduces the sense of reality by removing objects from the environment in real time~\cite{dr}. The removed regions can be filled with previously observed backgrounds~\cite{preobserve} or captured using multiple cameras~\cite{multi-camera}. 

Our study shares similarities with DR methods that use multiple cameras to capture the background. In our approach, the surgical field is recorded using cameras mounted on a surgical light. While methods like \cite{multi-camera} often depend on pre-calibrated cameras, the surgical lamp in our system may move during procedures, changing the cameras' relative positions. This requires a system that can automatically realign video frames.
\cite{obayashi} automatically performs camera calibration and homography estimation from manually selected video clips and utilizes them to make 3D video players. We fully automate the procedure of \cite{obayashi} and have developed a system that aligns the videos whenever cameras move.

\section{Method}
\label{sec:method}

\begin{figure}
    \centering
    \includegraphics[width=\linewidth]{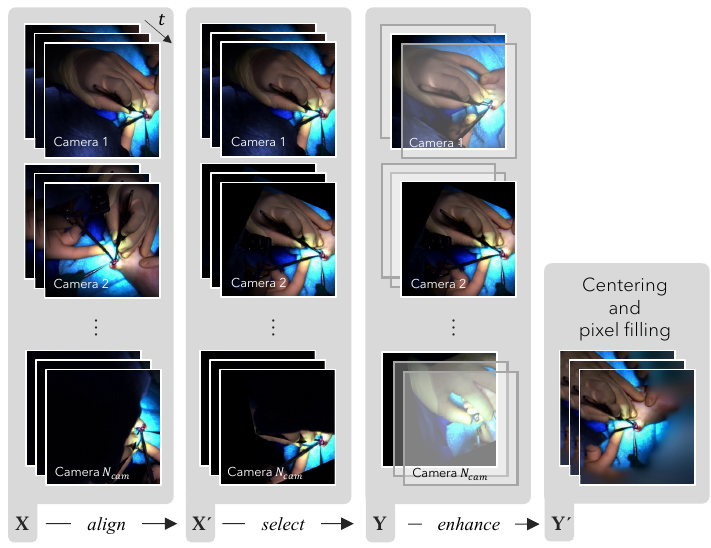}
    \caption{Overview of the proposed method. $\mathbf{X}$, $\mathbf{X}'$, $\mathbf{Y}$, and $\mathbf{Y}'$ denote all input, aligned, stabilized, and enhanced video frames, respectively.}
    \label{fig:entire_label}
\end{figure}

The proposed method consists of three major steps: view alignment, selection, and enhancement (\figurename~\ref{fig:entire_label}).
\begin{equation*}
    \mathbf{X} \xrightarrow[]{align} \mathbf{X}' \xrightarrow[]{select} \mathbf{Y} \xrightarrow[]{enchance} \mathbf{Y}',
\end{equation*}
where $\mathbf{X}$, $\mathbf{X}'$, $\mathbf{Y}$, and $\mathbf{Y}'$ denote all input frames, aligned frames, minimally occluded video frames, and enhanced video frames, respectively.
Consequently, our goal is to take $\mathbf{X}$ as input and produce $\mathbf{Y}'$ of minimally occluded single-view video.
\begin{itemize}
    \item \textit{Alignment:} We align all camera views, $\mathbf{X}$, to a representative one (e.g., $c=1$) and generate $\mathbf{X}'$.
    \item \textit{Selection:} We select video frames that observe the least occlusion at the moment among  $\mathbf{X}'$. We apply the learning-based object detection method by Shimizu \etal~\cite{Shimizu} for this purpose and compile the selected frames as a single video, $\mathbf{Y}$.
    \item \textit{Enhancement:} We enhance the output video quality. These enhancements involve bringing a surgical area of interest to the center and pixel compensation of disoccluded regions in $\mathbf{Y}$. The outcome is denoted as  $\mathbf{Y}'$.
\end{itemize}
Assuming $N_\text{cam}~(=5)$ cameras in McSL and $N_\text{img}$ recorded frames from the cameras,
$\mathbf{X} = \{X^c_t ~|~ c \in [1, N_\text{cam}],~t \in [1, N_\text{img}]\}$,
$\mathbf{X}' = \{X'^c_t ~|~ c \in [1, N_\text{cam}],~t \in [1, N_\text{img}]\}$,
$\mathbf{Y} = \{Y_t ~|~ t \in [1, N_\text{img}]\}$, and
$\mathbf{Y}' = \{Y'_t ~|~ t \in [1, N_\text{img}]\}$.
As such, $\mathbf{Y} \subset \mathbf{X}'$.
Details of our contributions (i.e., view alignment and enhancement processes) are provided in the following sections.

\subsection{View Alignment}
We align $N_\text{cam}$ views to a representative camera view (e.g., $c=1$) so that the view is stabilized over the frames. We achieve this by multi-camera calibration. That is, we calculate camera intrinsic and extrinsic parameters and a dominant plane in the scene to use it for homography warping of all views to the representative camera view.

\emph{Multi-Camera Calibration of Consecutive Frames:}
The number of distinctive visual features is often limited in surgical videos due to uniformly colored covers and hands (\fref{fig:mcsl}c), making feature detection and matching challenging with $N_\text{cam}$ frames at a given time point. We utilize the method by Obayashi \etal~\cite{obayashi}, which employs a neural network-powered feature point detector and matcher (i.e., Superpoint \cite{superpoint} and Superglue \cite{superglue}) over a sequence of frames to collect as many correspondences as possible for robust calibration. This approach is only valid for consecutive frames recorded by a non-moving McSL setup. Obayashi \etal~\cite{obayashi} propose manually identifying such a series of video frames using their dedicated video player.

\begin{figure}[t]
    \centering
    \includegraphics[width=\columnwidth]{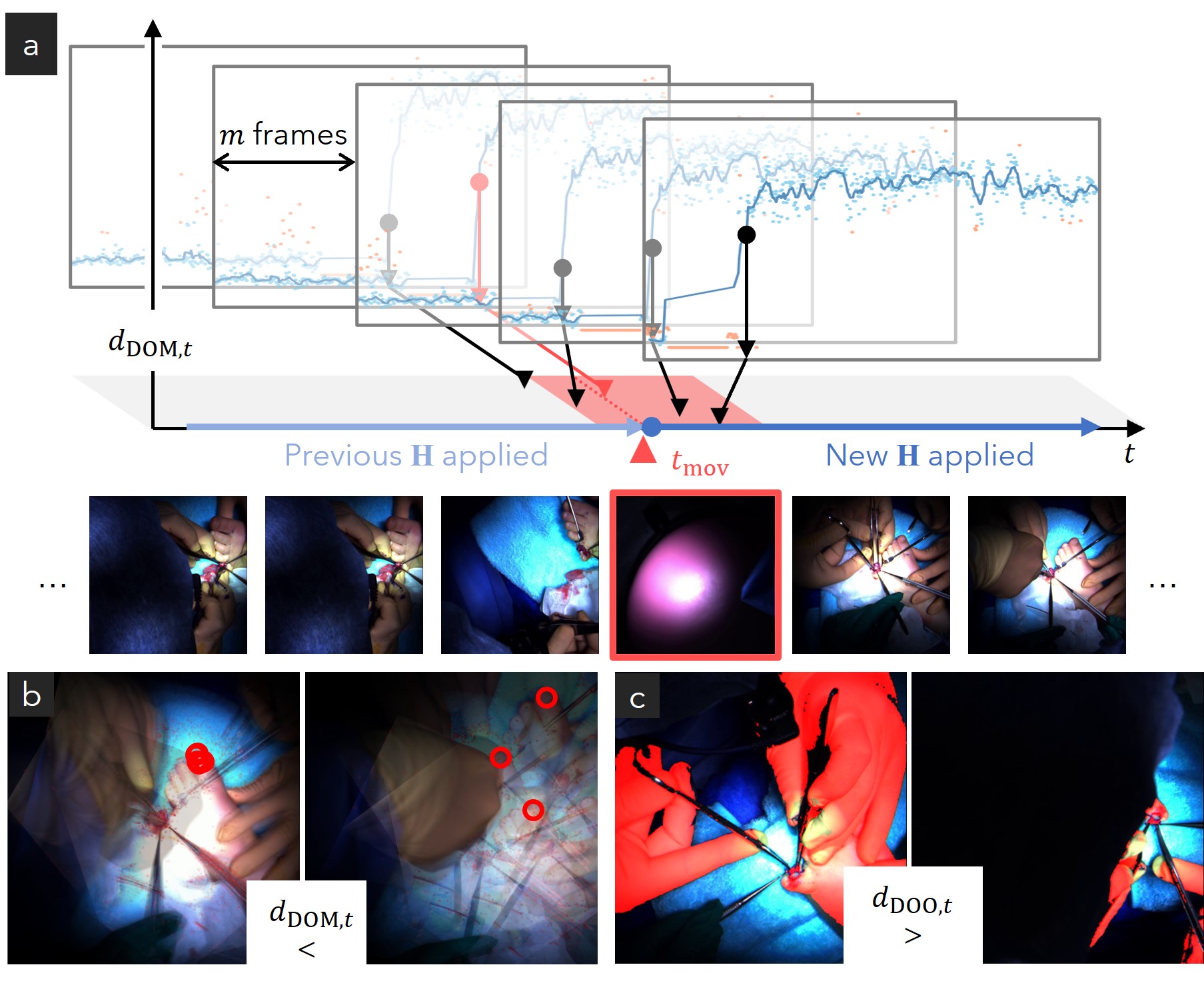}
    \caption{Robust McSL movement detection and homography estimation.
    (a) While the previous approach \citep{obayashi} asks the video viewer to find consecutive video frames to perform McSL calibration, our algorithm can detect such video frames automatically. We vote for detected frames in each bin of $m$ intervals and take the median as $t_\text{mov}$ as the outcome (i.e., McSL moved at $t_\text{mov}$).
    (b) The figures show the superimposed images of five cameras at (left) before and (right) after McSL moves. The red circles highlight example feature points of a common scene point (i.e., a big toe) to calculate the degree of misalignment, $d_{\text{DOM}, t}$, or moving McSL.
    (c) We distinguish surgical fields and the others in hue to collect non-occluded frames for stable homography calculation. The red highlights represent the detected surgical fields (i.e., $S_t$).
    }
    \label{fig:mcsl_motion_detection}
\end{figure}

We automate this calibration process by detecting time points when the McSL is moving over the frames, ensuring safe calibration of frames recorded with a non-moving McSL (\fref{fig:mcsl_motion_detection}). A common set of homography warping is applied to frames from the time the McSL stops moving to the time it starts moving again, resulting in $\mathbf{X}'$.

\emph{Camera Movement Detection:}
We identify sequences where McSL stays still to calculate geometric relationships between the $N_\text{cam}$ cameras. We align the static camera views with the calculated camera parameters until McSL moves at a frame index, $t_\text{mov}$.
To identify $t_\text{mov}$, we calculate the \textit{degree of misalignment}, $d_{\text{DOM}, t}$, which takes zero for a perfect alignment after a calibration or otherwise some larger values as the camera configuration changes (\fref{fig:mcsl_motion_detection}a and b). We calculate $d_{\text{DOM}, t}$ as an average projection error of all point correspondences among all $N_\text{cam}$ cameras at $t$.
\begin{equation}
    d_{\text{DOM}, t} = \frac{1}{N_\text{pts}}\sum^{N_\text{cam}}_c \sum^{N_\text{cam}}_{c'} \sum^{N^{c,c'}_\text{pts}}_i{\|H_{c\rightarrow c'}(\mathbf{p}^c_i) - \mathbf{p}^{c'}_i\|},
    \label{equ:degree of misalignment}
\end{equation}
where $\mathbf{p}^c_i$ is an $i_{th}$ SIFT \cite{SIFT} feature point in a camera $c$, $\mathbf{p}^{c'}$ is a corresponding SIFT feature point ($c \neq c'$), $H_{c\rightarrow c'}(\cdot)$ defines a homography warping from camera $c$ to $c'$, $N^{c,c'}_\text{pts}$ is the number of such correspondences between cameras $c$ and $c'$, and $N_\text{pts}$ is the total number of correspondences.

$d_{\text{DOM},t}$ may exhibit fluctuations over the frames due to feature point mismatches and varying numbers of matches.
To smooth the $d_{\text{DOM},t}$ values for a robust camera movement detection, we filter out outliers using the isolation forest algorithm \cite{isolationforest} and further smooth the inliers with a moving average (\fref{fig:mcsl_motion_detection}a).
Then, we identify frame indices where the smoothed $d_{\text{DOM},t}$ exceeds a threshold $\tau_\text{DOM}$.
\begin{equation}
    \tau_\text{DOM}=\min{\left(\max{(d_{\text{DOM}, t})}+1,~
            \frac{2}{N_\text{sub}}\sum_{t=1}^{N_\text{sub}}{d_{\text{DOM}, t}} \right)}.
    \label{equ:th}
\end{equation}
Here, the accuracy of $t_{\text{mov}}$ significantly affect visual quality of the generated video. Therefore, we use the following strategy to robustly estimate $t_{\text{mov}}$.

We observed that cameras typically move only once within a ten-minute period. Based on this, we set $N_{\text{sub}}$ to the number of frames in a ten-minute segment, regardless of the overall video length.
If the camera moves multiple times within a processing segment, the largest camera movement would determine the threshold and prevent robust detection of the other camera movements.
Limiting camera movements to a single instance within a processing segment can avoid this issue.
Therefore, we perform the above detection at every $m$ seconds ($m = 75$ by default) and find a frame index where $d_{\text{DOM}, t}$ exceeds $\tau$ more than or equal to four times within an $m$-seconds interval. Within the $m$ interval, different frame indices can be detected. Thus, we take their median as the final $t_\text{mov}$.

\emph{Occlusion-Free Frame Detection:}
We exclude frames that observe significant occlusions that lead to inaccurate homography calculations.
We automatically detect frame indices to calculate homography matrices, $t_\text{hom}$.

Frames right after $t_\text{mov}$, McSL still keeps moving until the surgeon finds its next location. While McSL moves, the surgeon grips it, and the arms occlude the camera field of view. Nonetheless, we should find a frame where no occlusion occurs for homography calculation. We evaluate the \textit{degree of occlusion}, $d_{\text{DOO}, t}$.
\begin{equation}
    d_{\text{DOO},t} = \left(\max{S_t} - \min{S_t}\right)/\bar{S_t},
\label{equ:s}
\end{equation}
where $S_t$ is an area of the surgical field, and $\bar{S_t}$ calculates the averaged $S_t$ over different $N_\text{cam}$ cameras.
Our intention is that all cameras observe the same area when nothing occludes the surgical field.
$S_t$ is distinguished in color space since, in surgery, areas of interest often have colors resembling skin and blood, while other areas, such as the surgeon's head and the operating table, are intended to contrast in hue.

We calculate $d_{\text{DOO},t}$ at every 30 frames (i.e., one second) and monitor whether the value exceeds $\tau_\text{DOO}$ (0.5 by default). When we observe $d_{\text{DOO},t}<\tau_\text{DOO}$ for $n$ times in a sequence, we select the frame indexes as $t_\text{hom}$.

\begin{figure}[tb]
    \begin{center}
    \includegraphics[width=\columnwidth]{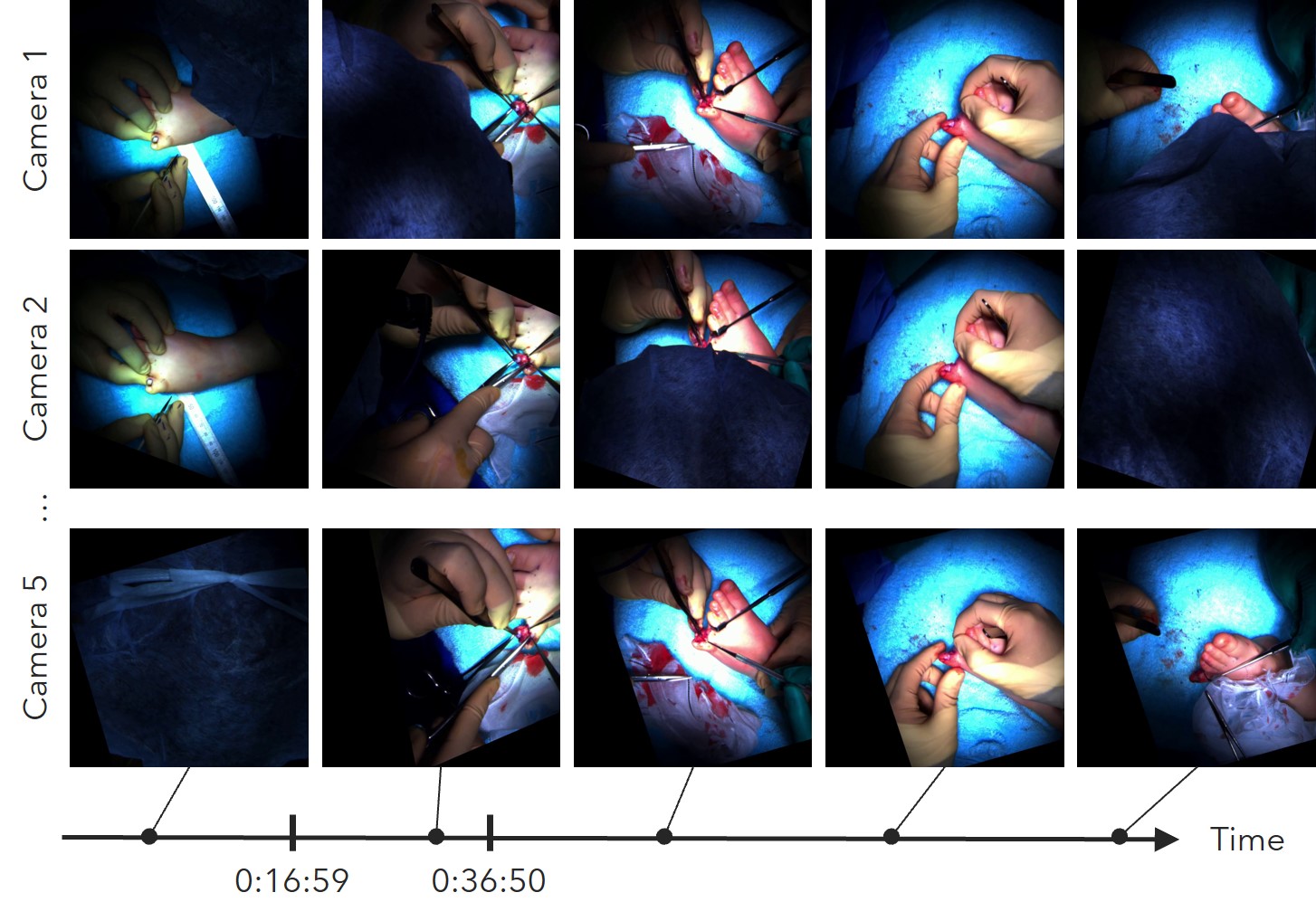}
    \caption{Auto-alignment with respect to camera 1 over time.
    The entire surgery lasts 1:39:19, and the times = 0:16:59 and 0:36:50 in the figure are the frame IDs detected by camera movement detection, indicating the moment when realignment becomes necessary.}
    \label{fig:auto-align}
    \end{center}
\end{figure}

\fref{fig:auto-align} shows the results of Auto-alignment using the proposed method. The figure also includes frames with detected camera movement $t_\text{mov}$ (\ie, frame ID at times = 0:16:59 and 0:36:50).
Once all five viewpoints were aligned, they were fed into the camera-switching algorithm to generate a virtual single-viewpoint video. 
The method requires about 40 minutes every time the cameras move. Please note that, unlike the existing manual alignment methods, it is fully automated and requires no human-involved interactions. 

\subsection{Visibility Improvement of Single-view Video}
\label{sec:enchancement}

We implemented two visual improvements to address artifacts in generated videos (\fref{fig:mcsl}c). Namely, we put the surgical field at the center of the screen and inpaint missing pixels in the synthesized videos. The video editing of these two methods can be toggled on demand according to the user's preferences.

\emph{Centering Surgical Field:}
The surgical field may be located around the edges of the video frame after warping.
To shift the area of interest back to the center of the video frame, we identify the surgical field.
To this end, we collect gaze locations of interests of five surgeons who participated in the expert review in \sref{sec:3}.
We used a commercial eye tracker and took the average of the five gaze locations in every frame.
We then shift frames with the collected offsets after the alignment and frame selection processes.
To secure all pixels of the original frame within the field of view after warping, we warp the original frame to a twice as large frame and then crop it to the original size.

\emph{Filling Missing Region:}
After Homography warping and shifting the frames, missing pixels appear (\fref{fig:mcsl}c).
Such pixels are represented as black pixels, and videos with large black parts are destructive. Therefore, we synthesize the other camera frames, $\mathbf{X}'$, warped to the view, $\mathbf{Y}$, to compensate for the missing pixels.
When aligning five camera views, the proposed method selects one camera ($c=1$ in this paper) as the reference viewpoint to align the other camera views. As such, there is always one camera involving no homography transformation and no missing pixels. We use images of this camera to fill in the missing pixels.

\begin{figure}[t]
    \begin{center}
    \includegraphics[width=1.0\hsize]{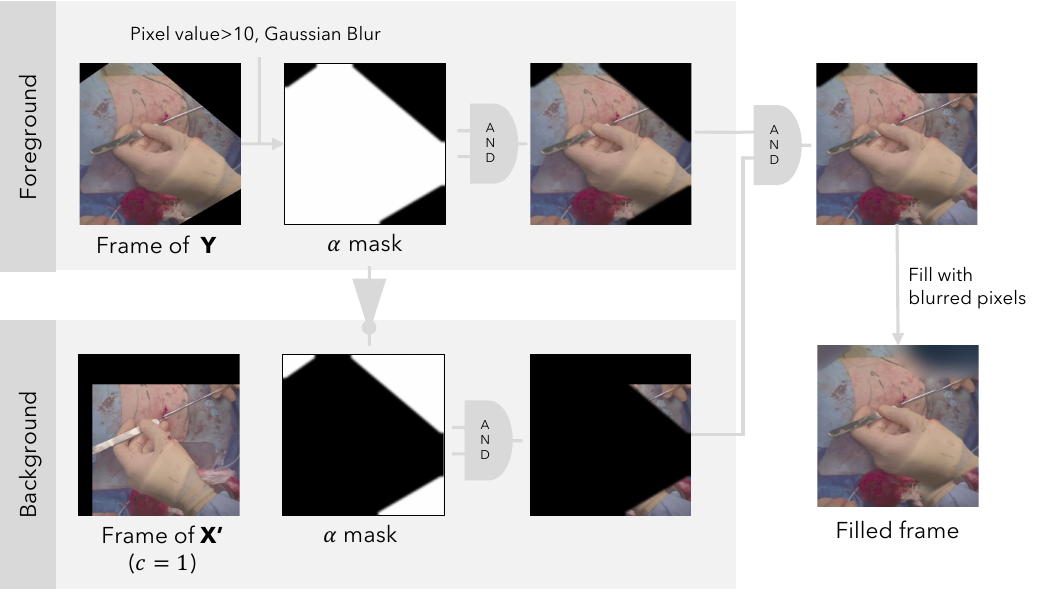}
    \caption{
        The procedure for combining two images during filling missing region. The foreground image is the video Y with missing pixels, and the background image is the video from the warp destination viewpoint (Ours is the Camera 1 viewpoint). First, regions in the foreground image with pixel values above 10 are extracted and blurred to generate an alpha mask. Using this mask, the foreground and background are blended to perform alpha blending at the boundaries. For pixels that remain missing after this process, the pixel values of the previous frame are blurred and retained.}
    \label{fig:pixelfilling}
    \end{center}
\end{figure}

However, homography warping for a planar surface does not approximate 3D objects well, especially at a close range. A resembled image shows clear boundaries between warped images.
Therefore, we apply alpha blending to blend warped images at a view seamlessly (\fref{fig:pixelfilling}).
We extract a binary image from an image at a representative view (i.e., $\mathbf{Y}$) that exceeds pixel values of $10$ and obtain an alpha mask by applying Gaussian blur with a kernel size of 49 pixels.
This alpha mask is then used to blend the foreground, $\mathbf{Y}$, with the background images from the other viewpoints, $\mathbf{X}'$.
Another point to consider is that the reference camera view may also shift and observe missing pixels during the centering process. These missing pixels are filled with pixels from previous frames, rather than through the processes mentioned above. Here, we assume these previous frames are already complete.
Since using temporally distant frames may result in misalignment of the hand or the surgical site, more weight is assigned to temporally closer frames by applying more blur to temporally distant frames during the synthesis process.
The synthesized image is output.

\begin{figure*}[tb]
    \begin{center}
    \includegraphics[width=\hsize]{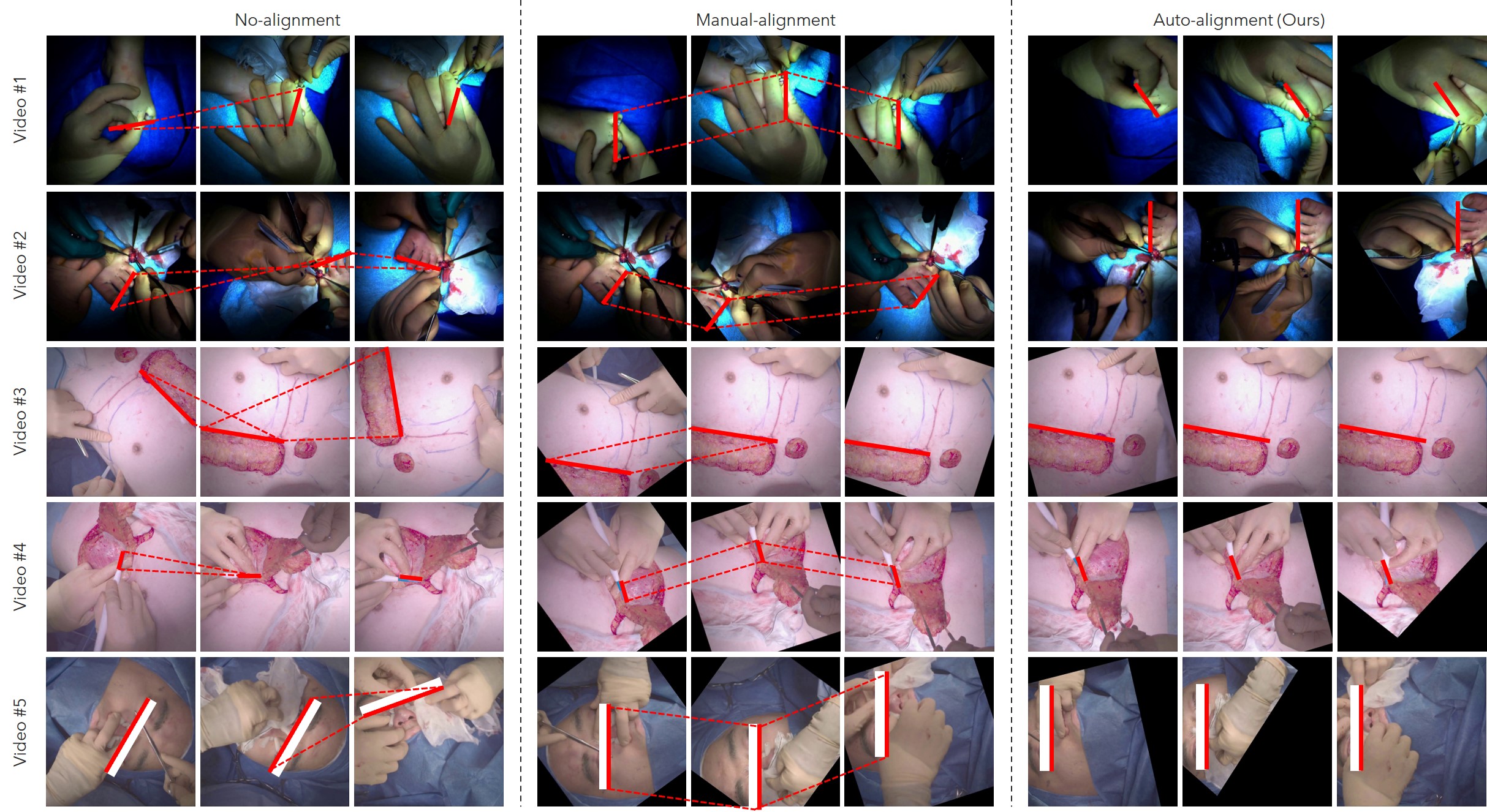}
    \caption{Example video frames from the three methods (No-alignment, Manual-alignment, and Auto-alignment). The red lines indicate positions and orientations of certain parallels.
    In the videos generated with No-alignment, the positions and orientations change every time the light (i.e., cameras) moves, which makes video observations difficult. Manual-alignment shows more stable results while it shows greater misalignment compared to ours. Ours shows reduced misalignments between viewpoints.}
    \label{fig:result}
    \end{center}
\end{figure*}

\section{Evaluations and Experts Review}
\label{sec:3}

We conducted experiments to investigate our method's efficacy. We first explain the implementation details in \sref{sec:31}. Then we compare our method with other baselines in \sref{sec:32}. Following these experiments, \sref{sec:33} investigates the effect of our visibility improvement approaches.

\subsection{Experimental Setups}\label{sec:31}

\emph{Dataset:} To capture our original dataset, we used a multi-camera system attached to a surgical light to capture videos of surgical procedures. We captured three types of actual surgical procedures: polysyndactyly, anterior thoracic keloid skin graft, and posttraumatic facial trauma rib cartilage graft. 
From these videos, we prepared five videos which were trimmed to one minute each. Video \#1 and \#2 show the surgery of polysyndactyly, video \#3 and \#4 show the anterior thoracic keloid skin graft scene, and video \#5 shows the surgery of posttraumatic facial trauma rib cartilage graft.

\emph{Hardware Setup:} We used Ubuntu 20.04 LTS OS, an Intel Core i9-12900 for the CPU, and 62GB of RAM. We defined the area of surgical field as hue ranging from 0 to 30 or from 150 to 179 in HSV color space in OpenCV. 

\emph{Evaluation Metrics:}
Our method aims to reduce the misalignment between viewpoints that occurs when swıtching between multiple cameras and generate single-view surgical videos with less occlusion.
Following a previous work that calculated degree of misalignment between consecutive time-series frames~\citep{evalindex2}, we used two metrics, the interframe transformation fidelity (ITF) and the average speed (AvSpeed).
ITF represents the average peak signal-to-noise ratio (PSNR) between frames as
\begin{equation}
\text{ITF}=\frac{1}{N_\text{img} - 1} \sum_{j=1}^{N_\text{img}-1} \text{PSNR}(I_j, I_{j+1})\text{,}
\label{equ:itf}
\end{equation}
where $N_\text{img}$ is the total number of frames and $I_j$ is the $j$th frame of the output video. ITF is higher for videos with less motion blur.
AvSpeed expresses the average speed of feature points. With the total number of frames $N_\text{img}$ and the number of all feature points in a frame $N_\text{p}$, AvSpeed is calculated as
\begin{equation}
\text{AvSpeed}=\frac{1}{N_\text{p}(N_\text{img} - 1)} \sum_{k=1}^{N_\text{p}}\sum_{j=1}^{N_\text{img}-1} \| z_{k}(j+1) - z_{k}(j) \|\text{,}
\label{equ:avspeed}
\end{equation}
where $z_{k}(j)$ denotes the image coordinates of the feature point. 

\subsection{Evaluating Alignment Approaches}\label{sec:32} 

To compare our method against baseline methods, we conducted a subjective evaluation. 11 physicians involved in surgical procedures regularly who were expected to actually use the surgical videos were selected as participants. 

\emph{Baselines:}
We compared our automatic alignment method (Auto-alignment) with two conventional methods. In one of these methods, which is used in a hospital camera switching is performed after manual alignment (Manual-alignment). The other method switches between camera views with no alignment (No-alignment).

\emph{Results:}
\fref{fig:result} shows single-view video frames generated by our method and the two conventional methods. 
The red lines indicate the position and orientation of the instep of the patient's foot. In the video generated of No-alignment, the position and orientation of the insteps changed every time the camera views were switched. This switching makes the surgical region harder to be watched. Manual-alignment presents better results than No-alignment but was characterized by greater misalignment than the proposed method. It should also be noted that manual alignment requires time and effort. Our method can effectively keep the misalignment to a minimum.

\begin{figure}[tb]
    \begin{center}
    \includegraphics[width=\columnwidth]{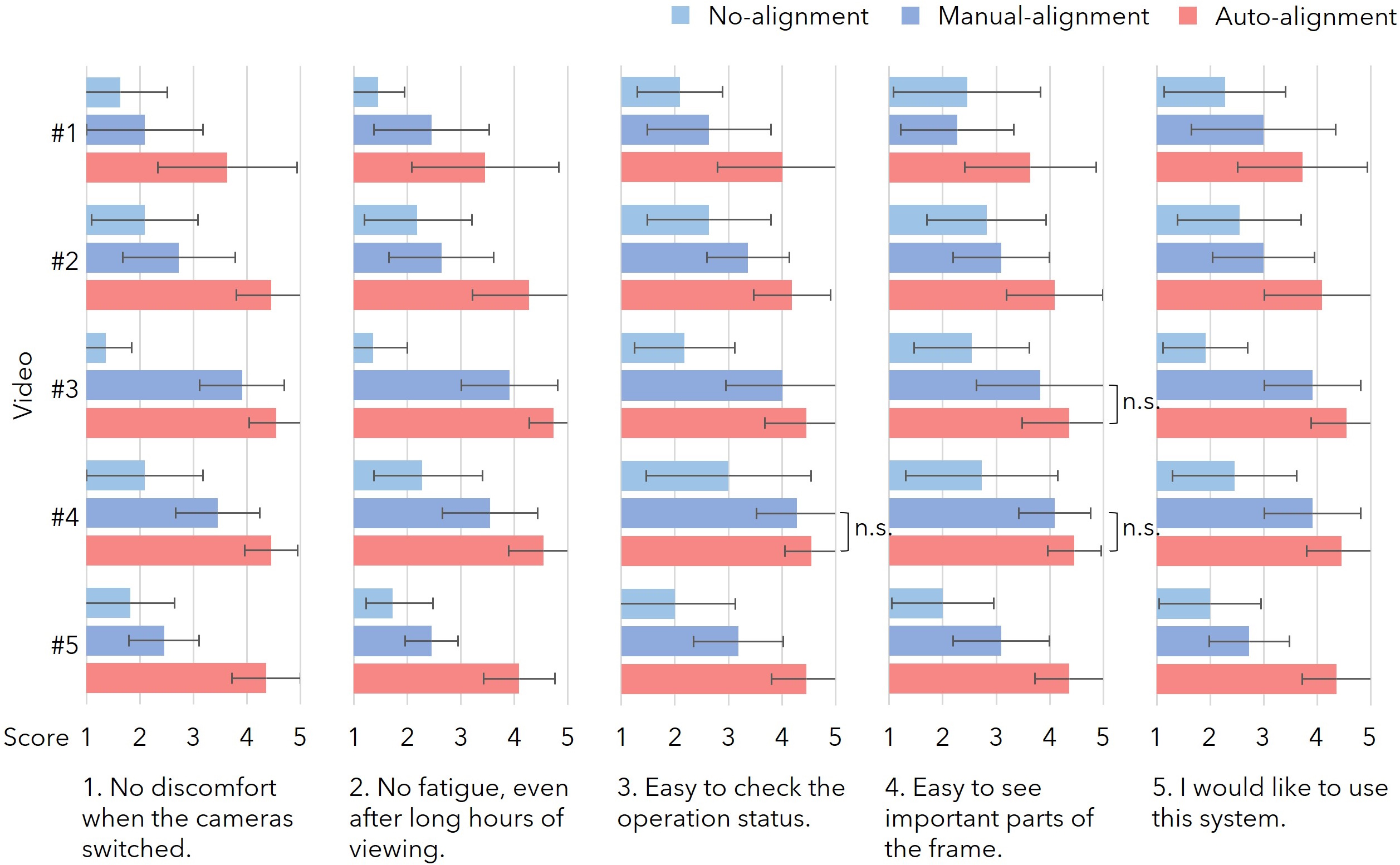}
    \caption{Results of the expert review. A nonparametric Friedman test showed significant differences in all videos and factors. In addition, a Wilcoxon signed-rank test was conducted to evaluate significant differences only between the previously used manual-alignment and the proposed auto-alignment method. We observed significant differences for almost all videos and factors except for Video \#3. n.s. indicates a no significant pair.}
    \label{fig:qualitative_evaluation}
    \end{center}
\end{figure}

\emph{Expert Review:}
Following a previous work~\citep{evalindex1}, we recruited eleven experienced surgeons who were expected to actually use surgical videos and collected their subjective evaluations to perform a qualitative comparison between the three methods. The participants were asked to score the videos on five points between 1 (``disagree'') and 5 (``agree'') for the following factors.
\begin{enumerate}[1.]
  \item No discomfort when the cameras switched.
  \item No fatigue, even after long hours of viewing.
  \item Easy to check the operation status.
  \item Easy to see important parts of the frame.
  \item I would like to use this system.
\end{enumerate}

\fref{fig:qualitative_evaluation} summarizes the results.
A nonparametric Friedman test showed significant differences in all videos and factors. However, since No-alignment was not actually used, significant differences only between Manual-alignment and Auto-alignment should be evaluated. 
Therefore, the Wilcoxon signed-rank test at 5\% significance level was also conducted between these two methods.
We observed significant differences between the proposed method (Auto-alignment) and the two conventional methods for almost all videos and factors except for Video \#3 - Statement 4 (p = 0.095) and Video \#4 - Statements 3 (p = 0.149) and 4 (p = 0.072).

The results of our superior scores in factors 1 and 2 suggest that our method generates a stable video with little misalignment between camera viewpoints. Similarly, the results in factors 3 and 4 suggest that the video generated by our method makes it easier to confirm the surgical area. Furthermore, as shown in factor 5, the proposed method received the highest score regarding the participants' willingness to use the system in actual medical practice. 
We believe that the proposed method can contribute to improving the quality of medical care by facilitating stable observation of the surgical field.

Although we observed statistically significant differences between the proposed method and the baselines for almost all the test videos, significant differences were not observed only in Video \#3, factor 3 and Video \#4, factors 3 and 4.
There could be two reasons for this. The small number of participants: Since we limited the participants to experienced surgeons, obtaining a larger sample size was quite difficult. Differences in surgical field geometries: Our method is more effective for scenes with a three-dimensional geometry. If the surgical field is flat with fewer feature points, as in the case of the anterior thoracic keloid skin graft procedure, differences between our method and the manual alignment mehod, which does not take into account three-dimensional structures, are less likely to be observed.

After the experiment, some surgeons pointed out that visibility could be improved if the surgical field was centered or black missing areas were less noticeable. This comment inspired the user-selectable functions in Sect. \ref{sec:enchancement}.

\begin{table}[tb]
\footnotesize
\centering
\caption{Results of the quantitative evaluation.}\label{tab:quantitative}
\scalebox{0.9}{ 
\begin{tabular}{cccccccc}
    \toprule
     & \multicolumn{3}{c}{ITF [dB] ($\uparrow$)} & & \multicolumn{3}{c}{AvSpeed [pixel/frame] ($\downarrow$)}\\
     \cmidrule{2-4} \cmidrule {6-8}
     & \multicolumn{3}{c}{Alignment} & & \multicolumn{3}{c}{Alignment}\\
     & No & Manual & Auto (Ours) & & No & Manual & Auto (Ours)\\
    \midrule
    \#1 & 11.97 & 11.87 & \textbf{17.54} & & 406.3 & 416.1 & \textbf{166.1}\\
    \#2 & 11.30 & 11.93 & \textbf{15.77} & & 339.4 & 328.7 & \textbf{195.6}\\
    \#3 & 16.17 & 17.85 & \textbf{22.26} & & 448.6 & 230.9 & \textbf{92.2}\\
    \#4 & 14.43 & 16.01 & \textbf{19.26} & & 379.0 & 240.7 & \textbf{77.5}\\
    \#5 & 15.19 & 17.42 & \textbf{21.66} & & 551.6 & 383.2 & \textbf{169.6}\\
    \bottomrule
    & &
\end{tabular}
}
\end{table}

\emph{Performance evaluation in reducing the misalignment:}
To investigate the performance in reducing the misalignment between viewpoints that occurs when swıtching between multiple cameras, we conducted a quantitative evaluation to assess the degree of misalignment between video frames based on ITF and AvSpeed that we introduced in Sect. \ref{sec:31}.

The results are shown in \tref{tab:quantitative}.
The ITF of the videos generated using the proposed method was 20\%-50\% higher than that of the videos with manual alignment. The AvSpeed of the videos generated using the proposed method was 40\%–70\% lower than that of the videos with manual alignment, indicating that the shake was substantially corrected.

\subsection{Evaluating Visual-Aid Approaches}\label{sec:33} 

We conducted subjective evaluation experiments to verify whether the implementations intended to improve the visibility of single-view videos actually enhance visibility.

\begin{figure*}[htb]
    \begin{center}
    \includegraphics[width=\hsize]{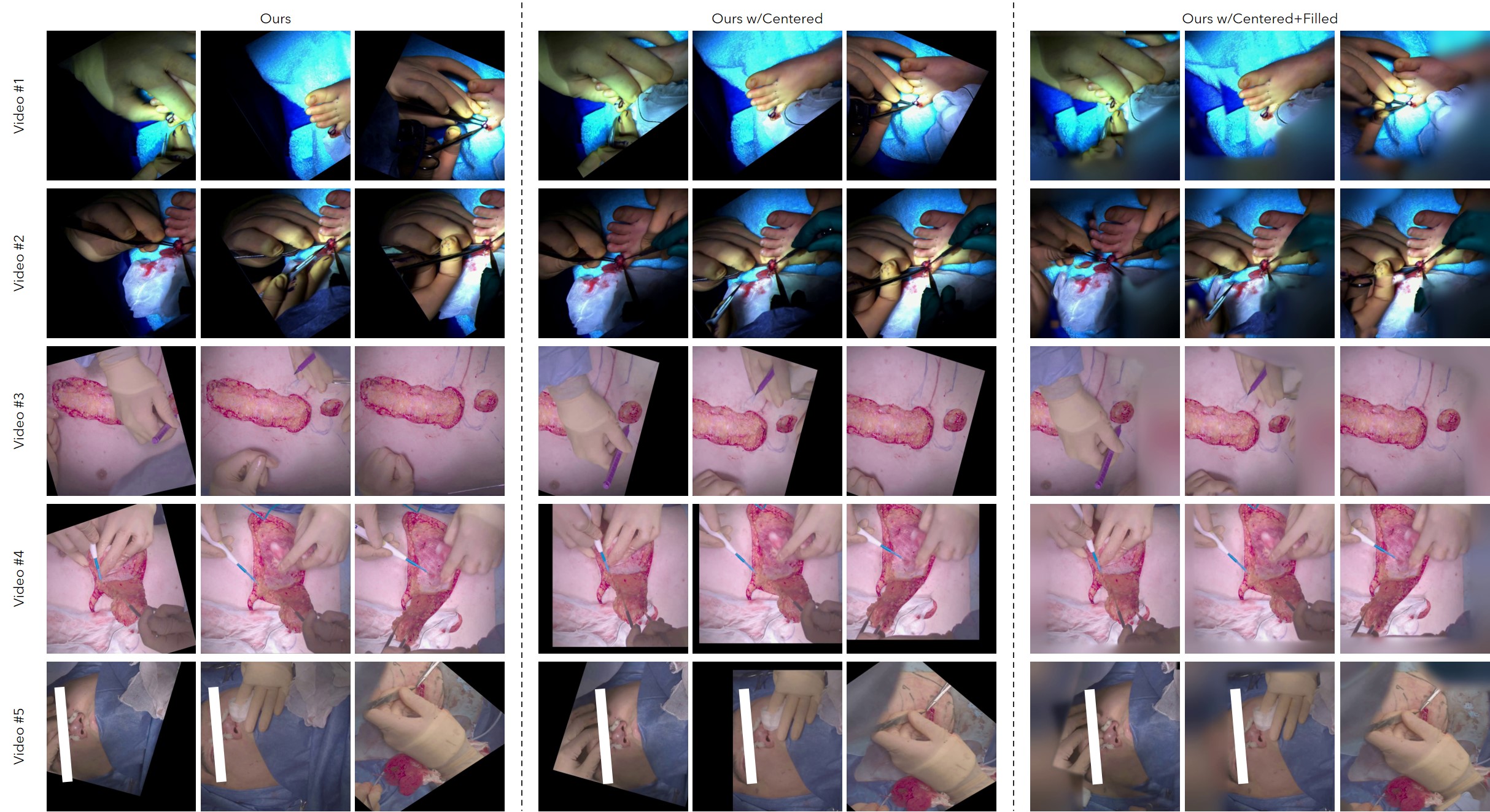}
    \caption{Example video frames from the three methods: Auto-alignment alone (Ours), Auto-alignment plus centering surgical field (Ours w/Centered), and Auto-alignment plus centering surgical field and filling missing region (Ours w/Centered+Filled). 
    In the video of Ours w/Centered, the surgical field that the doctor wants to see is centralized; in the image of Ours w/Centered+Filled, the missing areas of pixel values are complemented.}
    \label{fig:result_extension}
    \end{center}
\end{figure*}

\emph{Baselines:}
We compared our Auto-alignment method (Ours) with additional two methods incorporating enhancements aimed at improving generation quality as follows: Auto-alignment plus centering surgical field (Ours w/Centered), and Auto-alignment plus centering surgical field and filling missing region (Ours w/Centered+Filled). \fref{fig:result_extension} shows single-view video frames generated by our methods with the addition of Centering and Filling. 

\emph{Expert Review:}
11 physicians were selected as participants, including people different from \ref{sec:32}.
The participants were asked to score four factors, 1 (``disagree'') to 5 (``agree'').
\begin{enumerate}[1.]
  \item No discomfort at the moment the cameras switch.
  \item No fatigue or stress when viewing.
  \item Easy to see the surgical field.
  \item I want to actually use such video.
\end{enumerate}

\begin{figure}[h!]
    \begin{center}
    \includegraphics[width=\hsize]{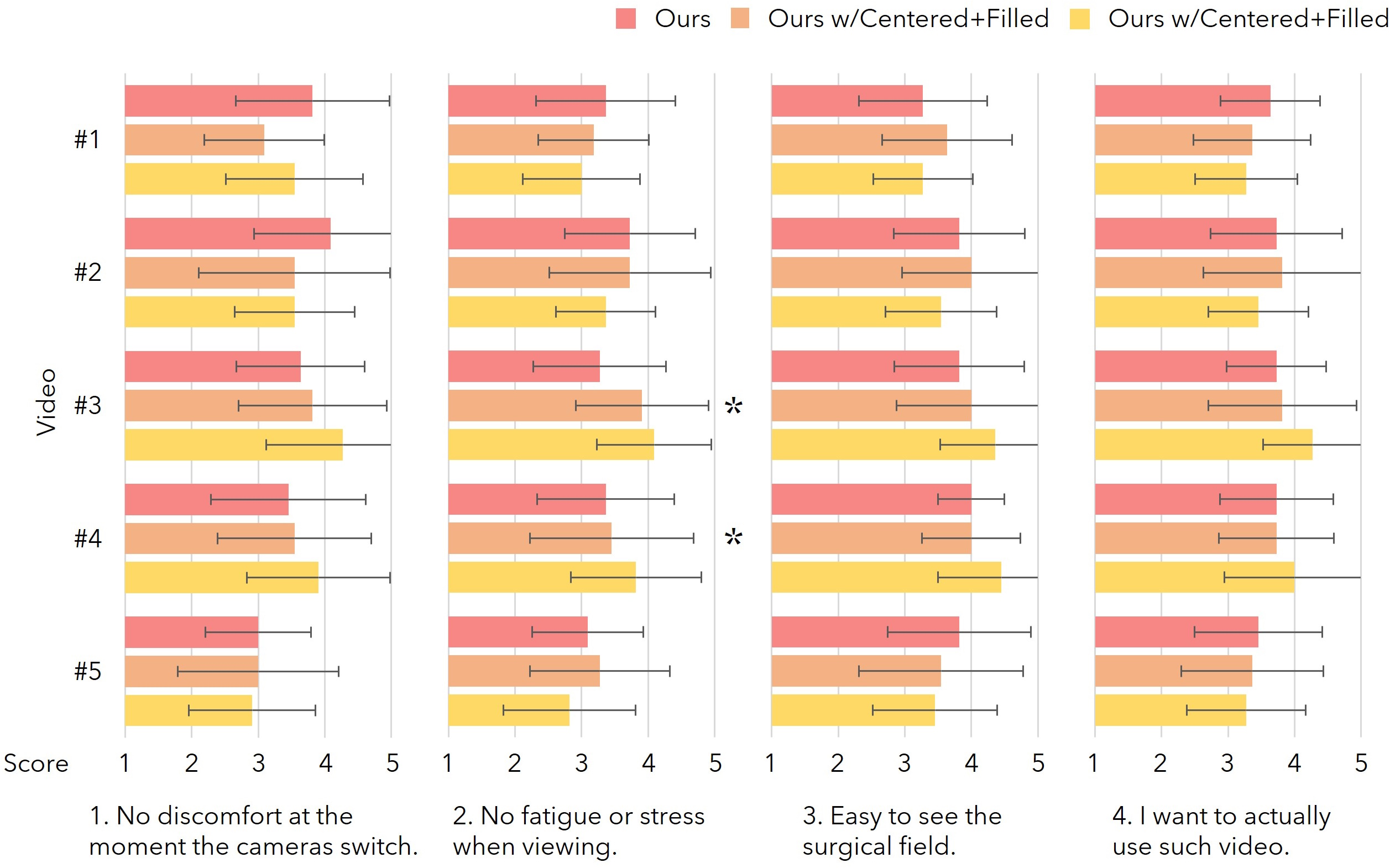}
    \caption{Results of the subjective evaluation experiment of proposed methods with several functions. The Friedman test confirmed significant differences only in factor 2 of Video \#3 and \#4. Almost all the videos and factors did not show significantly higher (or lower) scores for Ours w/Centered or Ours w/Centered+Filled. * indicates a significant difference.}
    \label{fig:subjective_evaluation}
    \end{center}
\end{figure}

\fref{fig:subjective_evaluation} shows the results.
The Friedman test confirmed significant differences only in factor 2 of Video \#3. Almost all the videos and factors did not show significantly higher (or lower) scores for Ours w/Centered or Ours w/Centered+Filled.
The figure shows that the preferred method tends to differ depending on the type of surgery. (Polysyndactyly in Video \#1 and \#2 has a higher score of Ours, anterior thoracic keloid skin graft in Video \#3 and \#4 has a higher score of Ours w/Centered+Filled, etc.).
We also received comments on the experiment from the participants and found individual differences in whether participants felt these synthesis approaches affected the visibility of single-view video.
Therefore, we believe it would be beneficial to allow users to choose whether to use or not to use the visual aids, centering, and filling according to their individual preferences.

\section{Conclusion and Discussion}

We proposed a method for generating virtual single-viewpoint surgical videos using multiple cameras attached to a shadowless lamp without occlusion or misalignment through automatic geometric calibration.
In evaluation experiments, we compared our Auto-alignment method with Manual-alignment and No-alignment. The results verified the superiority of the proposed method both qualitatively and quantitatively. 
Additionally, we conducted a survey of surgeons' preferences for the two video edits performed for Auto-alignment videos. Based on their feedback, we proposed design guidelines for video synthesis.
The ability to easily confirm the surgical field with the automatically generated virtual single-viewpoint surgical video will contribute to medical treatment.

\emph{Limitations:}
Our method relies on visual information to detect the timing of homography calculations (i.e., $t_\text{hom}$). However, we may use prior knowledge of a geometric constraint such that cameras are at the pentagon corners (\figurename~\ref{fig:entire_label}).
We assume that the McSL does not move more than once in ten minutes for a robust calculation of $d_\text{mov}^t$. Although surgeons rarely moved the light more often, fine-tuning the parameter may result in further performance improvement. The current implementation shows misaligned images if the cameras move more frequently.
In the user-involved study, several participants reported noticeable black regions where no camera views were projected. (e.g., \fref{fig:result}). One possible complement is to project pixels from other views.

\section*{Acknowledgments}
\noindent
\textbf{Ethical approval} Approval for open surgery video recording was obtained from the ethics committee of Keio University under 20180111.

\noindent
\textbf{Consent to participate} Informed consent was obtained from all individual participants included in the study.

\noindent
\textbf{Funding}
This work was partially supported by JSPS KAKENHI [Grant Number 22H03617] and the Alexander von Humboldt Foundation funded by the German Federal Ministry of Education and Research.

\bibliographystyle{IEEEtran}
\bibliography{main}

\end{document}